\documentclass[10pt,twocolumn,letterpaper]{article}

\usepackage{iccv}
\usepackage{times}
\usepackage{epsfig}
\usepackage{graphicx}
\usepackage{amsmath}
\usepackage{amssymb}
\usepackage{color}
\usepackage{times}
\usepackage{epsfig}
\usepackage{algorithm}
\usepackage{algorithmicx}
\usepackage{algpseudocode}
\usepackage{graphics}
\usepackage{threeparttable}
\usepackage{color}
\usepackage[normalem]{ulem}
\usepackage{multirow}
\usepackage{float}
\usepackage{amsfonts}
\usepackage{bm}
\usepackage{subfig}
\usepackage{array}
\usepackage[table]{xcolor}
\usepackage{colortbl}
\usepackage{pifont}
\usepackage{cite}
\usepackage{diagbox}
\usepackage{rotating}
\usepackage{booktabs}
\usepackage{overpic}
\usepackage{textcomp}
\usepackage{contour}

\ifdefined \GramaCheck
  \newcommand{\CheckRmv}[1]{}
  \newcommand{\figref}[1]{Figure 1}%
  \newcommand{\tabref}[1]{Table 1}%
  \newcommand{\secref}[1]{Section 1}
  \newcommand{\eqnref}[1]{Equation 1}
\else
  \newcommand{\CheckRmv}[1]{#1}
  \newcommand{\figref}[1]{Fig.~\ref{#1}}%
  \newcommand{\tabref}[1]{Tab.~\ref{#1}}%
  \newcommand{\secref}[1]{Sec.~\ref{#1}}
  \newcommand{\eqnref}[1]{Eq.~(\ref{#1})}
\fi

\definecolor{bblue}{rgb}{0,150,230}
\definecolor{mygray}{gray}{.92}

\makeatletter
\newcommand{\thickhline}{%
    \noalign {\ifnum 0=`}\fi \hrule height 1pt
    \futurelet \reserved@a \@xhline
}

\newcommand{\ConfInf}{\vspace{-.7in} {\normalsize \normalfont \color{blue}{
   IEEE International Conference on Computer Vision (ICCV) 2019}} \vspace{.45in} \\}

\makeatother

\newcommand{\supp}[1]{\textcolor{magenta}{#1}}

\def\ourmeasure{Scoot}

\usepackage[pagebackref=false,breaklinks=true,letterpaper=true,colorlinks,bookmarks=false]{hyperref}

\iccvfinalcopy 

\def\iccvPaperID{1527} 

\ificcvfinal\fi

\graphicspath{{./Imgs/}}
\DeclareGraphicsExtensions{.jpg,.pdf,.png}
\begin{document}

\title{\ConfInf Scoot: A Perceptual Metric for Facial Sketches}

\author{Deng-Ping Fan$^{1,2}$ \quad
        ShengChuan Zhang$^4$ \quad
        Yu-Huan Wu $^1$ \quad
        Yun Liu $^1$ \quad \\
        Ming-Ming Cheng$^{1,}$\thanks{M.M. Cheng (cmm@nankai.edu.cn) is the corresponding author.}\quad 
        Bo Ren$^1$ \quad
        Paul L. Rosin$^3$ \quad
        Rongrong Ji$^{4,5}$ \\
    \small{$^1$} \small TKLNDST, CS, Nankai University \quad
    \small{$^2$} \small Inception Institute of Artificial Intelligence (IIAI) \quad
    \small{$^3$}  \small Cardiff University \\
    \small{$^4$} \small Department of Artificial Intelligence, School of Informatics, Xiamen University \quad
    \small{$^5$} \small Peng Cheng Lab\\
    {\tt \small \supp{http://mmcheng.net/scoot/}}
}

\maketitle
\thispagestyle{empty}

\begin{abstract}
\vspace{-8pt}
Human visual system has the strong ability to quick assess 
the perceptual similarity between two facial sketches. 
However, existing two widely-used facial sketch metrics, \eg, 
FSIM and SSIM fail to address this perceptual similarity in 
this field. Recent study in facial modeling area has verified
that the inclusion of both structure and texture has a significant
positive benefit for face sketch synthesis (FSS).
%
%
%
%
%
But which statistics are more important, 
and are helpful for their success?
In this paper, we design a perceptual metric, called
Structure Co-Occurrence Texture (\textbf{Scoot}),
which simultaneously considers the block-level spatial structure
and co-occurrence texture statistics.
To test the quality of metrics, we propose three novel meta-measures
based on various reliable properties.
Extensive experiments demonstrate that our Scoot metric exceeds the 
performance of prior work.
Besides, we built the first large scale (152k judgments)
human-perception-based sketch database that can evaluate how
well a metric is consistent with human perception.
%
%
%
Our results suggest that ``spatial structure'' and ``co-occurrence texture''
are two generally applicable perceptual features in face sketch synthesis.
\end{abstract}

\vspace{-8pt}
\section{Introduction}\label{sec:Introduction}
\vspace{-4pt}
Comparison of data is considered to be the most important step\cite{zhang2018unreasonable,Zhao2019Edge},
especially in image processing~\cite{zhao2018flic,yizhe_zsl_2017,cheng2019bing}.
%
For various end-user applications such as face sketch~\cite{tang2003face},
image style transfer~\cite{isola2017image},
image quality assessment~\cite{wang2004image}, saliency detection
\cite{fan2019shifting,fan2018salient,fan2019D3Net,zhao2019Contrast},
segmentation~\cite{shen2016real,shen2014lazy,shen2018submodular} and
disease classification~\cite{yang2018clinical}, image denoising~\cite{Xu_2017_ICCV},
the comparison can turn out to be evaluating a ``perceptual distance'', which assesses
how similar two images are in a way that highly correlates with \emph{human perception}.

In this paper, we study facial sketch and show that human judgments are often
different from current evaluation metrics, and as the first related attempt,
we provide a novel perceptual distance for sketch according to human choice principles.
%
As noticed in~\cite{zhang2018unreasonable},
human judgments of similarity depend on high-order image
structure. Facial sketches are made up of a lot of textures,
and there are many algorithms for synthesizing sketches,
which is a good fit for this problem.
However, designing a good perceptual metric should
take into account human perception in facial sketch comparison, which
should:
\begin{itemize}
\vspace{-5pt}
\item  closely match \textbf{human perception} so that
good sketches can be directly used in various
subjective applications, \eg, law enforcement and entertainment.

\vspace{-5pt}
\item be \textbf{insensitive to slight mismatches} (\ie, re-size, rotation)
  since real-world sketches drawn by artists do not precisely match each
  pixel to the original photos.

\vspace{-5pt}
\item be capable of \textbf{capturing holistic content},
that is, prefer the complete sketch to one that only
contains strokes (\ie, has lost some facial components).
\end{itemize}

\CheckRmv{
\begin{figure}[t!]
\small
\centering
    \begin{overpic}[width=\columnwidth]{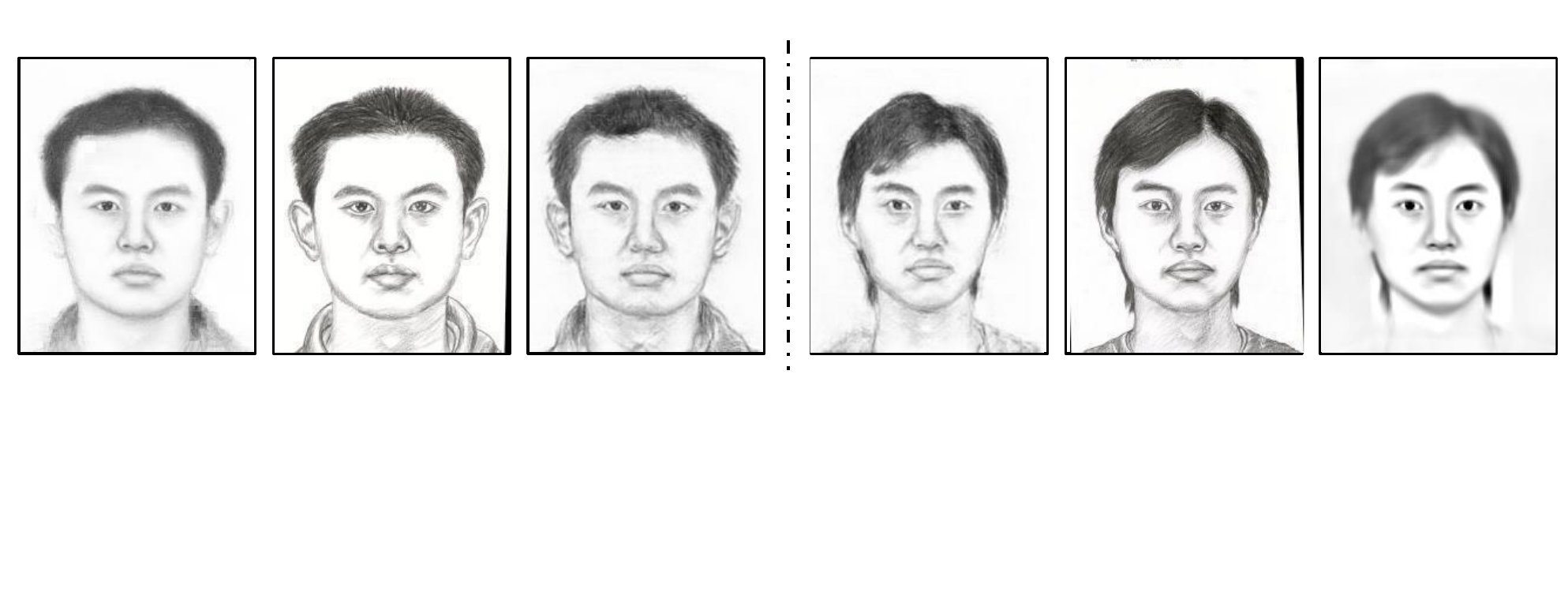}
    \put(6,36){S0}
    \put(23,36){R}
    \put(40,36){S1}

    \put(58,36){S0}
    \put(75,36){R}
    \put(89,36){S1}

    \put(35,11){Humans}
    \put(4,5.5){Others}
    \put(37,0){Scoot}
    
    \put(53,11){Humans}
    \put(87,5.5){Others}
    \put(55,0){Scoot}
    
    \end{overpic}
    \vspace{-10pt}
    \caption{\small 
    \textbf{Which synthesized sketch is more similar to the middle sketch?}
    For the right case, sketch 0 (S0) is more similar than sketch 1 (S1) \textit{w.r.t.} reference (R) in terms of structure and texture.
    Sketch 1 almost completely destroys the texture of the hair.
    The widely-used (SSIM~\cite{wang2004image}, FSIM~\cite{zhang2011fsim}), classic
    (IFC~\cite{sheikh2005information}, VIF~\cite{sheikh2006image}) and 
    recently released (GMSD~\cite{xue2014gradient})
    metrics disagree with humans. Only our Scoot metric agrees well with humans. 
    }\label{fig:existingFailureCase}
\end{figure}
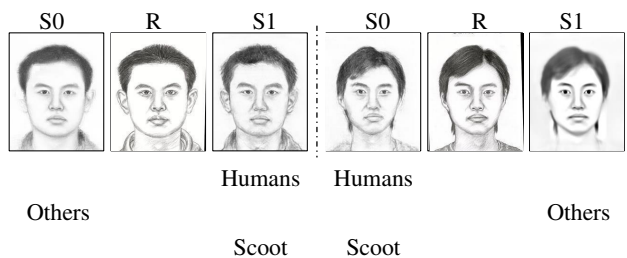
}

\CheckRmv{
\begin{table*}[t!]
  \centering
  \footnotesize
  \renewcommand{\arraystretch}{1.0}
  \renewcommand{\tabcolsep}{1.8mm}
  \begin{tabular}{l|rl|c|c|r|l|rl|c|c|r}
  \hline\thickhline
  \rowcolor{mygray}
   No. & \textbf{Model} & \textbf{Year'Pub}&\textbf{Sj.}& \textbf{Rr.} & \textbf{Ob.} & No. & \textbf{Model} & \textbf{Year'Pub}&\textbf{Sj.}& \textbf{Rr.} & \textbf{Ob.}\\
  \midrule
  \midrule
  1& \textbf{ST} \cite{tang2003face}       & 03'ICCV  &  & VRR  & & 2 & \textbf{STM} \cite{tang2004face}        & 04'TCSVT &  &VRR&\\
  3& \textbf{LLE} \cite{liu2005nonlinear}   & 05'CVPR  &  & VRR & & 4 & \textbf{BTI} \cite{liu2007bayesian}    & 07'IJCAI &  & &RMSE\\
  5& \textbf{E-HMMI} \cite{gao2008local}      & 08'NC    &  & VRR & UIQI & 6 & \textbf{EHMM} \cite{gao2008face}       & 08'TCSVT &  & VRR & \\
  7& \textbf{MRF} \cite{wang2009face}       & 08'PAMI  &  & VRR & & 8& \textbf{SL} \cite{xiao2010photo}       & 10'NC    &  & VRR & UIQI\\
  9& \textbf{RMRF} \cite{zhang2010lighting} & 10'ECCV  &  & VRR & &10& \textbf{SNS-SRE} \cite{gao2012face}       & 12'TCSVT &  & VRR & \\
  11& \textbf{MWF} \cite{zhou2012markov}    & 12'CVPR  &  & VRR & &12& \textbf{SCDL} \cite{wang2012semi}     & 12'CVPR  &  & & PSNR\\
  13& \textbf{Trans} \cite{wang2013transductive} & 13'TNNLS  &   & VRR & & 14& \textbf{SFS-SVR} \cite{wang2013heterogeneous}  & 13'PRL  &  & VRR & VIF\\
  15& \textbf{Survey} \cite{wang2014comprehensive} & 14'IJCV  & & &  RMSE, UIQI, SSIM &16& \textbf{SSD} \cite{song2014real}      & 14'ECCV & SV & VRR & \\
  17& \textbf{SFS} \cite{zhang2015face}    & 15'TIP  &  & VRR & FSIM, SSIM &18& \textbf{FCN} \cite{zhang2015end}      & 15'ICMR & ES & VRR &\\
  19& \textbf{RFSSS} \cite{zhang2016robust}   & 16'TIP  &  & VRR & FSIM, SSIM &20& \textbf{KD-Tree} \cite{zhang2016fast} & 16'ECCV  &  & VRR & VIF, SSIM\\
  21& \textbf{MrFSPS} \cite{peng2016multiple}& 16'TNNLS &  & VRR & FSIM, VIF, SSIM &22& \textbf{2DDCM} \cite{tu2016facial}    & 16'TIP  &  & VRR & FSIM, SSIM \\
  23& \textbf{RR} \cite{wang2017data}       & 17'NC   &  & VRR & VIF, SSIM& 24& \textbf{Bayesian} \cite{wang2017bayesian}& 17'TIP &  & VRR & VIF, SSIM\\
  25& \textbf{RFSSS} \cite{zhang2017face}    & 17'TCSVT&  &  VRR & FSIM, SSIM &26& \textbf{S-FSPS} \cite{peng2017superpixel}& 17'TCSVT &   &  VRR & FSIM, VIF, SSIM\\
  27& \textbf{ArFSPS} \cite{li2017adaptive} & 17'NC &   & VRR & FSIM &28& \textbf{BFCN} \cite{zhang2017content} & 17'TIP & SV &  VRR & \\
  29& \textbf{DGFL} \cite{zhu2017deep}      & 17'IJCAI &   & VRR & SSIM &30& \textbf{FreeH} \cite{li2017free}      & 17'IJCV & SV & &\\
  31& \textbf{Pix2pix} \cite{isola2017image}    & 17'CVPR  &  & & &32& \textbf{CA-GAN} \cite{gao2017composition}& 17'CVPR &  &VRR &  SSIM\\
  33& \textbf{ESSFA} \cite{fivser2017example}& 17'TOG &   &  &     &34& \textbf{PS$^2$MAN} \cite{wang2018high}& 18'FG &  & VRR &FSIM, SSIM\\
  35& \textbf{NST} \cite{semmo2017neural}   & 17'NPAR &  &  &&36& \textbf{CMSG} \cite{zhang2018compositional}& 18'TC & SV & VRR & \\
  37& \textbf{RSLCR} \cite{wang2018random}  & 18'PR &  & VRR &SSIM& 38& \textbf{MRNF} \cite{zhang2018markov}  & 18'IJCAI &  &  & VIF, SSIM\\
  39& \textbf{$\rho$GAN} \cite{zhang2018robust}& 18'IJCAI &  &  &FSIM& 40& \textbf{FSSN} \cite{jiao2018modified} & 18'PR &  &  &PSNR, SSIM \\
  41& \textbf{MAL} \cite{zhang2018face}     & 18'TNNLS &  &   &FSIM, SSIM& 42& \textbf{MRNF} \cite{zhang2018coarse}   & 18'AAAI &  & VRR & VIF, SSIM\\
  \hline\thickhline
  \end{tabular}
  \vspace{-10pt}
  \caption{\small
  \textbf{Summarization of 42 representative FSS-based  algorithms.}
  \textbf{Sj.}: Subjective metric.
  \textbf{Rr.}: Recognition rates.
  \textbf{Ob.}: Objective metric.
  SV = Subjective Voting. ES = Empirical Study.
  VRR = various recognition rate methods, such as, null-space LDA~\cite{chen2000new},
  Random Sampling LDA~\cite{wang2004random,wang2006random}, dual-space LDA~\cite{wang2004dual}, LPP~\cite{he2004locality}, Sparse Representation and Classification~\cite{wright2009robust}.
  Note that UIQI~\cite{wang2002universal} is a special case of SSIM~\cite{wang2004image}.
  }\label{tab:MetricSummary}
  \vspace{-10pt}
\end{table*}
}

To the best of our knowledge, no prior metric can satisfy
all these properties simultaneously.

For example, in face sketch synthesis (FSS),
the target is for the synthetic sketch to be indistinguishable
from the reference by a human subject, 
although their pixel representations might be mismatched.
Let us take a look at \figref{fig:existingFailureCase} in which
there are three examples. Which one is closer to the middle reference?
While this comparison task seems trivial for humans, to date the
widely-used metrics disagree with human judgments.
Not only are visual patterns very high-dimensional, 
but the very notion of visual similarity is often
subjective~\cite{zhang2018unreasonable}.




%


Our contributions to the facial sketch community can be summarized in three points.
Firstly, as described in \secref{sec:ProposedAlgorithm},
we propose a Structure Co-Occurrence
Texture (\textbf{Scoot}) perceptual metric for FSS
that provides a unified evaluation considering both structure and texture.

Secondly, as described in \secref{sec:Meta-measures}, we design three
meta-measures based on the above three reliable properties. Extensive experiments on these meta-measures verify that
our \ourmeasure~metric exceeds the performance of prior works.
Our experiments indicate that ``spatial structure'' and ``co-occurrence''
texture are two generally applicable perceptual features in FSS.

Thirdly, we explore different ways of exploiting  texture statistics (\eg, Gabor, Sobel, and Canny, \etc).
We find that simple texture features~\cite{galloway1974texture,gabor1946theory}
performs far better than the commonly used metrics in the literature~\cite{sheikh2005information,
wang2004image,zhang2011fsim,sheikh2006image,xue2014gradient}.
Based on our findings, we construct the first large-scale human-perception-based
sketch database that can evaluate how well a metric is in line with human perception.
%

Our three contributions presented above offer a complete metric benchmark suite,
which provides a novel view and a practical tool (\eg, metric, meta-measures and database)
to analyze data similarity from the human perception direction.

\CheckRmv{
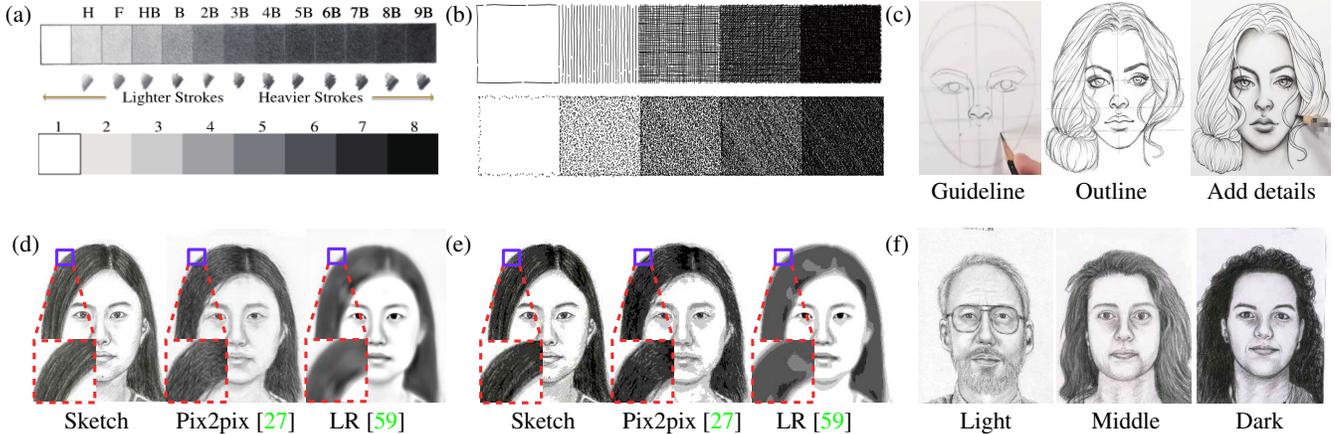
\begin{figure*}[t!]
\small
\centering
    \begin{overpic}[width=\textwidth]{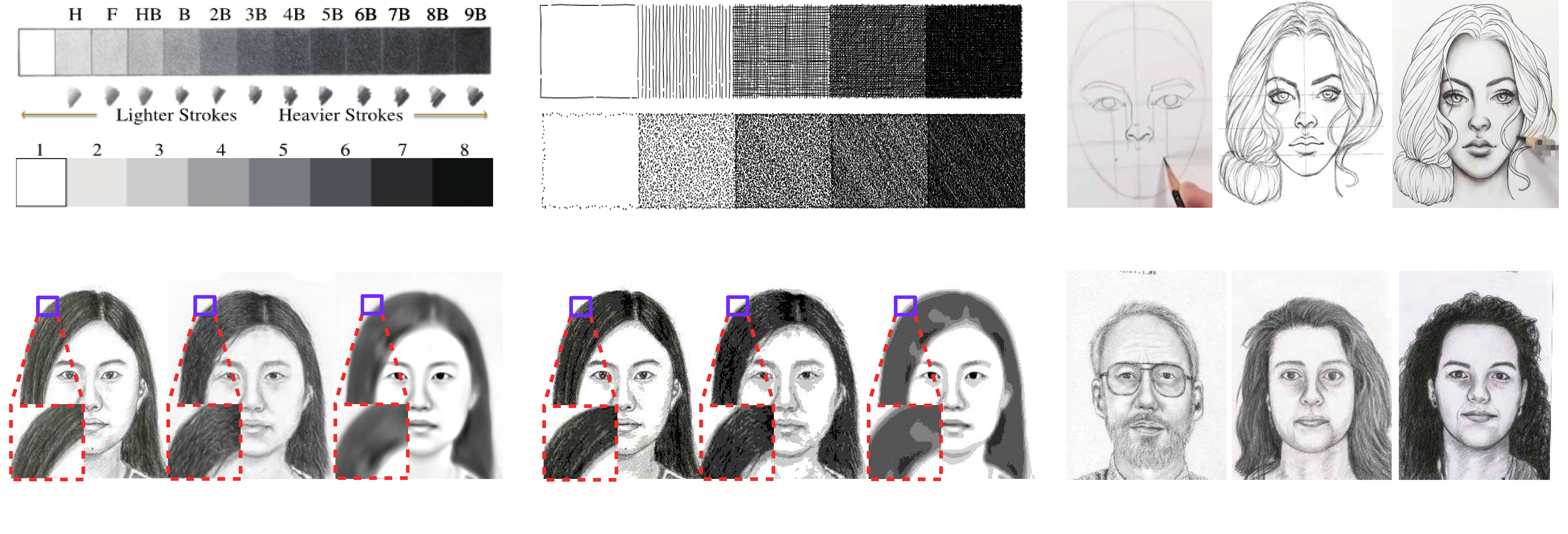}
    \put(-1.5,32.5){(a)}
    \put(32,32.5){(b)}

    \put(65.5,32.5){(c)}
    \put(69,19){Guideline}
    \put(80,19){Outline}
    \put(90,19){Add details}

    \put(-1,15){(d)}
    \put(3,1.5){Sketch}
    \put(11.5,1.5){Pix2pix~\cite{isola2017image}}
    \put(23.2,1.5){LR~\cite{wang2017data}}

    \put(37,1.5){Sketch}
    \put(45.5,1.5){Pix2pix~\cite{isola2017image}}
    \put(57.2,1.5){LR~\cite{wang2017data}}


    \put(32,15){(e)}

    \put(65.5,15){(f)}
    \put(71.2,1.5){Light}
    \put(81.2,1.5){Middle}
    \put(92.2,1.5){Dark}

    \end{overpic}
    \vspace{-20pt}
    \caption{\small \textbf{Motivation of the proposed \ourmeasure~metric.}
    (a) Pencil grades and their strokes.
    (b) Using stroke tones to present texture. The stroke textures used,
    from top to bottom, are: ``cross-hatching'', ``stippling''. The stroke
    attributes, from left to right, are: spare to dense. Images are from~\cite{winkenbach1994computer}.
    (c) The artist draws the sketch from guideline to details.
    (d) The original sketches.
    (e) The quantized sketches.
    (f) Creating various tones of the stroke by applying different pressure (\eg, light to dark) on the pencil tip.
    }\label{fig:SecondImage}
    \vspace{-10pt}
\end{figure*}
}

\vspace{-4pt}
\section{Related Work}
\vspace{-4pt}
From~\tabref{tab:MetricSummary}, we observe that some works utilize
recognition rates (Rr.) to 
evaluate the quality of synthetic sketches. However, Rr. cannot completely
reflect the visual quality of synthetic sketches~\cite{wang2016evaluation}.
In the FSS area, the widely-used perceptual metrics, \eg, SSIM~\cite{wang2004image},
FSIM~\cite{zhang2011fsim}, and VIF~\cite{sheikh2006image} were initially
designed for image quality assessment (IQA) which aims 
to evaluate image distortion such as Gaussian blur, jpeg, and jpeg 2000 compression.
Directly introducing the IQA metric to FSS may be
intractable
(see \figref{fig:existingFailureCase}) due to the different nature of their task.

Psychophysics~\cite{zucker1989two} and prior work, \eg, line
drawings~\cite{grabli2004programmable,freeman2003learning}
indicate that human perception of sketch similarity depends on two crucial
factors, \ie, image \emph{structure}~\cite{wang2004image} and \emph{texture}~\cite{wang2016evaluation}.
However, how perceptual are these so-called ``perceptual features''?
Which elements are critical for their success?
How well do these ``perceptual features'' actually correspond to human visual perceptions?
As noticed by Wang \etal~\cite{wang2016evaluation},
there is currently no reliable perceptual metric in FSS.
We review the topics most pertinent to facial sketch within
the constraints of space:

\textbf{Heuristic-based Metric.}
The most widely used metric in FSS is SSIM proposed by Wang \etal~\cite{wang2004image}.
SSIM computes structure similarity, and luminance and contrast comparison using a sliding window on the local patch.
Sheikh and Bovik~\cite{sheikh2006image} proposed the VIF metric which evaluates the
image quality by quantifying two kinds of information.
One is obtained via the human visual system channel, with the input
ground truth and the output reference image information.
The other is achieved via the distortion channel, called distortion information, and the result is the ratio of these two types of information.
%
%
Studies of the human vision system (HVS) found that the features perceived by human vision are consistent
with the phase of the Fourier series at different frequencies.
Therefore, Zhang \etal~\cite{zhang2011fsim} chose phase congruency as the primary feature.
Then they proposed a low-level feature similarity metric called FSIM.

Recently, Xue~\etal~\cite{xue2014gradient} devised a simple metric named
gradient magnitude similarity deviation (GMSD), where the pixel-wise
gradient magnitude similarity is utilized to obtain image local quality.
The standard deviation of the overall gradient magnitude similarity map is calculated as the final image quality index.
Their metric achieves the state-of-the-art (SOTA) performance
compared with the other metrics.

\textbf{Learning based Metric.}
As well as the heuristic-based metric, there are numerous
learning based metrics~\cite{dosovitskiy2016generating,gao2017deepsim,talebi2018nima},
for comparing images in a perceptual-based
manner which have been used to evaluate image compression and
many other imaging tasks.
We refer readers to a recent survey~\cite{zhang2018unreasonable} for a comprehensive
review of various deep features adopted for perceptual metrics.
This paper focuses on showing why face sketches require a specific perceptual
distance metric that differs from or improves upon previously heuristic-based methods.

\vspace{-4pt}
\subsection{Motivation}\label{sec:Motivation}
\vspace{-4pt}
We observed the basic
principles of the sketch and noted that
``graphite pencil grades'' and ``pencil's strokes'' are the two fundamental
elements in the sketch.

\vspace{-4pt}
\subsection{Graphite Pencil Grades.}
\vspace{-4pt}
In the European system, ``H'' \& ``B'' stand
for ``hard'' \& ``soft'' pencil, respectively.
\figref{fig:SecondImage}(a) illustrates the grade of graphite pencil.
Sketch images are expressed through a limited medium (graphite pencil)
which provides no color. Illustrator Sylwia Bomba~\cite{Sylwia2015Sketching}
said that ``if you put your hand closer to the end of the pencil,
you have darker markings. Gripping further up the
pencil will result in lighter markings.''
Besides, after a long period of practice, artists will form their
fixed pressure (\eg, from guideline to detail in \figref{fig:SecondImage}(c)) style.
In other words, the marking of the stroke can be varied (\eg, light to dark in \figref{fig:SecondImage}(f))
by changing the pressure on the pencil tip.
%
Note that different pressures on the tip will
result in various types of marking which is one of the quantifiable
factors called gray tone.

\emph{Gray Tone.} The quantification of gray tone should reduce the effect of slight amounts of noise and
over-sensitivity to subtle gray tone gradient changes in sketches. We introduce intensity quantization during the evaluation of gray tone similarity.
Inspired by previous works~\cite{clausi2002analysis},
we can quantize the input sketch $I$ to $N_{l}$ different
grades to reduce the number of intensities  to be considered:
$I'  = \Omega(I)$.
A typical example of such quantization is shown in \figref{fig:SecondImage}(d, e).
Humans will consistently rank Pix2pix higher than LR before (\figref{fig:SecondImage}(d))
and after (\figref{fig:SecondImage}(e)) quantizing the input sketches
when evaluating the perceptual similarity.
Although quantization may introduce artifacts, our experiments (\secref{sec:AblationStudy}) also show that this process can reduce sensitivity to minor intensity variations and balance
the performance and computational complexity.


\vspace{-4pt}
\subsection{Pencil's Strokes.}
\vspace{-4pt}
Because all of the sketches are generated by moving a tip on the paper, different paths of the tip along the paper will create various
stroke shapes. One example is shown in \figref{fig:SecondImage}(b),
in which different spatial distributions of the stroke have produced various textures
(\eg, sparse or dense). Thus, the stroke tone is another quantifiable factor.

\emph{Stroke Tone.}
The stroke tone and grey tone are not independent concepts. The gray tone is based on the different strokes of gray-scale
in a sketch image, while the stroke tone can be defined as the spatial distribution of gray tones.

An example is shown in \figref{fig:SecondImage}(d). Intuitively,
Pix2pix~\cite{isola2017image} is better than LR~\cite{wang2017data}
since Pix2pix preserves the texture (or stroke tone) of the hair and details in the face.
However, LR presents an overly smooth result and has lost much of the sketch style.

\vspace{-4pt}
\section{Proposed Algorithm}\label{sec:ProposedAlgorithm}
\vspace{-4pt}
This section explains the proposed \ourmeasure~metric, which
captures the \emph{co-occurrence texture} statistics in the
``block-level'' \emph{spatial structure}.
%
\vspace{-4pt}
\subsection{Co-Occurrence Texture}\label{sec:co-occurrenceTexture}
\vspace{-4pt}
With the two quantifiable factors at hand, we start to describe the details.
To simultaneously extract statistics about the ``stroke tone''
and their relationship to the surrounding ``gray tone'', we need
to characterize their \emph{spatial interrelationships}.
Previous work in texture~\cite{haralick1973textural} verified that
the \emph{co-occurrence matrix} can efficiently capture the
texture feature, due to the use of various powerful statistics.
Since the sketches show a lot of similarities to textures, we use
the co-occurrence matrix as our gray tone and stroke tone extractor.
%
%
Specifically, this matrix $\mathcal{M}$ is defined as:
\vspace{-8pt}
\begin{equation}\label{equ:HaralickMatrix}
\mathcal{M}_{(i,j)|d}=\sum_{y=1}^{H}\sum_{x=1}^{W}
\begin{cases}
1,\mbox{ if } I'_{x,y}=i \ \mbox{and} \ I'_{(x,y)+d}=j \\
0,\mbox{ otherwise}
\end{cases}
\vspace{-2pt}
\end{equation}
where $i$ and $j$ denote the gray value; $d=(\Delta x, \Delta y)$ is the relative distance
to $(x, y)$; $x$ and $y$ are the spatial positions in the given quantized sketch $I'$;
$I'_{x,y}$ denotes the gray value of $I'$ at position $(x,y)$;
$W$ and $H$ are the width and height of the sketch $I'$, respectively.
To extract the perceptual features in a sketch, we test the three most
widely used~\cite{haralick1979statistical} statistics:
Homogeneity ($\mathcal{H}$), Contrast ($\mathcal{C}$), and Energy ($\mathcal{E}$).

\emph{Homogeneity} reflects how much the texture changes in local regions,
it will be high if the gray tone of each pixel pair is similar.
The homogeneity is  defined  as:
\vspace{-8pt}
\begin{equation}\label{equ:Homogeneity}
    \mathcal{H} = \sum_{j=1}^{N_l}\sum_{i=1}^{N_l}{\frac{\mathcal{M}_{(i,j)|d}}{1+|i-j|}},
    \vspace{-5pt}
\end{equation}

\emph{Contrast} represents the difference between a pixel in $I'$ and its neighbor summed over
the whole sketch. This reflects that a low-contrast sketch is not characterized by
low gray tones but rather by low spatial frequencies. The contrast is highly correlated with
spatial frequencies. The contrast equals 0 for a constant tone sketch.
\vspace{-8pt}
\begin{equation}\label{equ:Contrast}
    \mathcal{C} = \sum_{j=1}^{N_l}\sum_{i=1}^{N_l}{|i-j|^2\mathcal{M}_{(i,j)|d}}
    \vspace{-5pt}
\end{equation}

\emph{Energy} measures textural uniformity. When only similar gray tones of pixels occur in
a sketch ($I'$) patch, a few elements in $\mathcal{M}$ will be close to 1, while others will be
close to 0.  Energy will reach the maximum if there is only one gray tone in a sketch ($I'$) patch. Thus, high energy corresponds to
the sketch's gray tone distribution having either a periodic or constant form.
\vspace{-8pt}
\begin{equation}\label{equ:Uniformity}
    \mathcal{E} = \sum_{j=1}^{N_l}\sum_{i=1}^{N_l}{(\mathcal{M}_{(i,j)|d})^2}
    \vspace{-5pt}
\end{equation}

\CheckRmv{
\begin{table}[t!]
  \centering
  \small
  \renewcommand{\arraystretch}{1.0}
  \setlength\tabcolsep{2pt}
  \resizebox{\columnwidth}{!}{
  \begin{tabular}{l}
  \hline\thickhline
  \textbf{Algorithm 1:} Structure Co-Occurrence Texture Measure\\
  \hline
  \textbf{Input:} Synthetic Sketch $X$, Ground Truth Sketch $Y$ \\
  Step 1: Quantize $X$ and $Y$ into $N_l$ grades \\
  Step 2: Calculate the matrices $\mathcal{M}(X)$ and  $\mathcal{M}(Y)$  according to Eq. 1\\
  Step 3: Divide the whole sketch image into a $k \times k$ grid of $k^2$ blocks \\
  Step 4: Extract the $\mathcal{CE}$ features according to Eq. 3 \& 4 from each block\\
  and concatenate them together\\
  Step 5: Compute the average feature of four orientations with Eq. 5\\
  Step 6: Evaluate the similarity between $X$ and $Y$ according to Eq. 6\\
  \textbf{Output:} Scoot score; \\
  \hline\thickhline
  \end{tabular}
  }
  \vspace{-10pt}
\end{table}
}

\vspace{-4pt}
\subsection{Spatial Structure}\label{sec:spatialStructure}
\vspace{-4pt}
To holistically represent the spatial structure, we follow the spatial
envelope strategy~\cite{oliva2001modeling,fan2017structure} to extract
the statistics from the ``block-level'' \emph{spatial structure} in the sketch.
First, we divide the whole sketch image into a $k \times k$ grid of $k^2$ blocks.
Our experiments demonstrate that the process can help to derive content information.
Second, we compute the co-occurrence matrix $\mathcal{M}$ for all blocks and
normalize each matrix such that the sum of its components is 1.
Then, we concatenate $p$ statistics (\eg, $\mathcal{H}, \mathcal{C}, \mathcal{E}$)
of all the $k^2$ blocks into a vector
\begin{math}
  \overrightarrow{\Phi}(I'_{s}|d)\!\in\!\mathbb{R}^{p\!\times\! k \!\times\! k}.
\end{math}

Note that each of the above statistics is based on a single direction
(\eg, $90^o$, that is $d$ = (0, 1)), since the direction
of the spatial distribution is also very
important to capture the style such as ``hair direction'',
``the direction of shadowing strokes''. To exploit this observation
for efficiently extracting the \emph{stroke direction} style, we compute
the average feature $\overrightarrow{\Psi}(I'_{s})$ of $T$ orientation
vectors to capture more directional information:
\vspace{-8pt}
\begin{equation}\label{equ:Direction}
    \overrightarrow{\Psi}(I'_{s})= \frac{1}{T}\sum_{i=1}^{T}{\overrightarrow{\Phi}}(I'_{s}|d_{i}),
    \vspace{-5pt}
\end{equation}
where $d_{i}$ denotes the $i$th direction and $\overrightarrow{\Psi}(I'_{s})\!\in\!\mathbb{R}^{p\!\times\! k \!\times\! k}$.

\CheckRmv{
\begin{figure*}[t!]
\small
\centering
    \begin{overpic}[width=\textwidth]{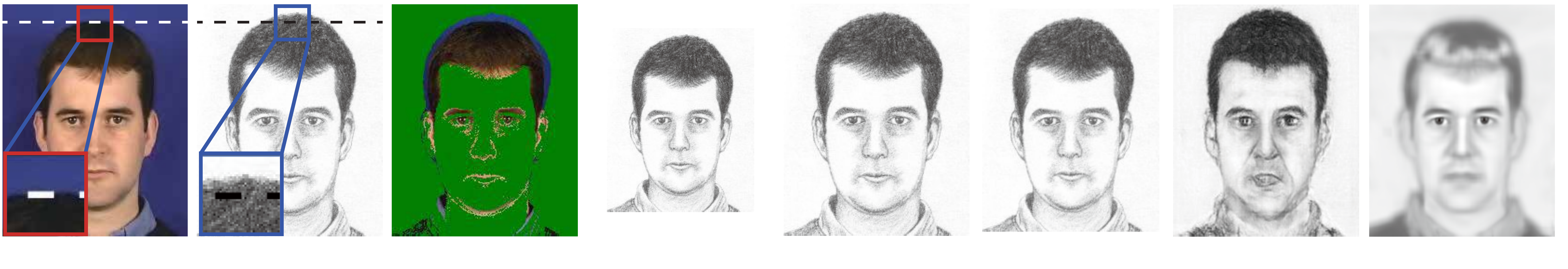}
    \put(2.7,0){(a) Photo}
    \put(13,0){(b) Reference}
	\put(25.4,0){(c) Difference}
    \put(37.3,0){(d) Downsized}

    \put(50.5,0){(e) Reference}
    \put(64,0){(f) Re-sized}
    \put(77,0){(g) Pix2pix}
    \put(91,0){(h) LR}
    \end{overpic}
    \vspace{-15pt}
    \caption{\small
    \textbf{Meta-measure 1: Stability to Slight Re-sizing.}
    }\label{fig:MM1}
    \vspace{-10pt}
\end{figure*}
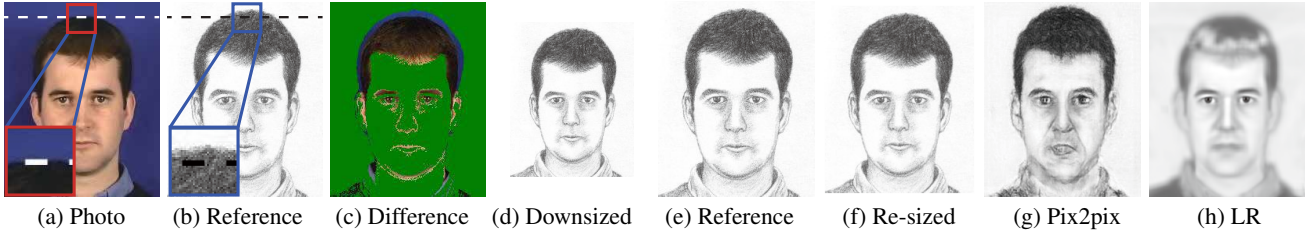
}

\vspace{-4pt}
\subsection{\ourmeasure~Metric}
\vspace{-4pt}
After obtaining the perceptual feature vectors of the reference sketch $Y$ and synthetic sketch $X$,
a function is needed to evaluate their similarity.
We have tested various forms of functions such as Euclidean distance or exponential functions, \etc,
but have found that the simple Euclidean distance is a simple and effective function and works
best in our experiments.
Thus, the proposed perceptual similarity \ourmeasure~metric can be defined as:
\begin{equation}\label{equ:Scoot}
\vspace{-8pt}
    E_{s} = \frac{1}{1+\left\|\overrightarrow{\Psi}(X'_{s})-\overrightarrow{\Psi}(Y'_{s})\right\|_2}.
\end{equation}
where $\left\|\cdot\right\|_2$ denotes the $l_2$-norm.
$X'_{s},~Y'_{s}$ denote the
quanitized
$X_{s},~Y_{s}$, respectively. $E_s=1$ represents identical style.

\vspace{-4pt}
\section{Experiments}
\vspace{-4pt}

\subsection{Implementation Details}
\vspace{-4pt}
The size of spatial structure $k$ in \secref{sec:spatialStructure}
is set to 4 to achieve the best performance.
The quantization parameter $N_{l}$ in \eqnref{equ:Scoot} is set to 6 grades.
We have demonstrated that $p$ = 2 (\eg, $\mathcal{C}$ in Eq. \ref{equ:Contrast}
combined with $\mathcal{E}$ in Eq. \ref{equ:Uniformity}) achieve the best
performance (see \secref{sec:AblationStudy}).
Due to the symmetry of the co-occurrence matrix $\mathcal{M}(i,j)$,
the statistical features in 4 orientations are actually equivalent
to the 8 neighbor directions at 1 distance. Empirically,
we set $T=4$ orientations $d_{i}\in \{(0,1), (-1,1), (-1,0), (-1,-1)\}$
to achieve the robust performance.

\subsection{Meta-measures}\label{sec:Meta-measures}
\vspace{-4pt}
As described in~\cite{margolin2014evaluate}, one of the most challenging
tasks in designing a metric is proving its performance.
Following~\cite{pont2013measures}, we use the
\emph{meta-measure} methodology, which is a measure that assesses a metric.
Inspired by~\cite{fan2017structure,Fan2018Enhanced,margolin2014evaluate},
we further propose three meta-measures based on the 3 properties described in \secref{sec:Introduction}.

\CheckRmv{
\begin{figure}[t!]
\small
\centering	
    \begin{overpic}[width=.92\columnwidth]{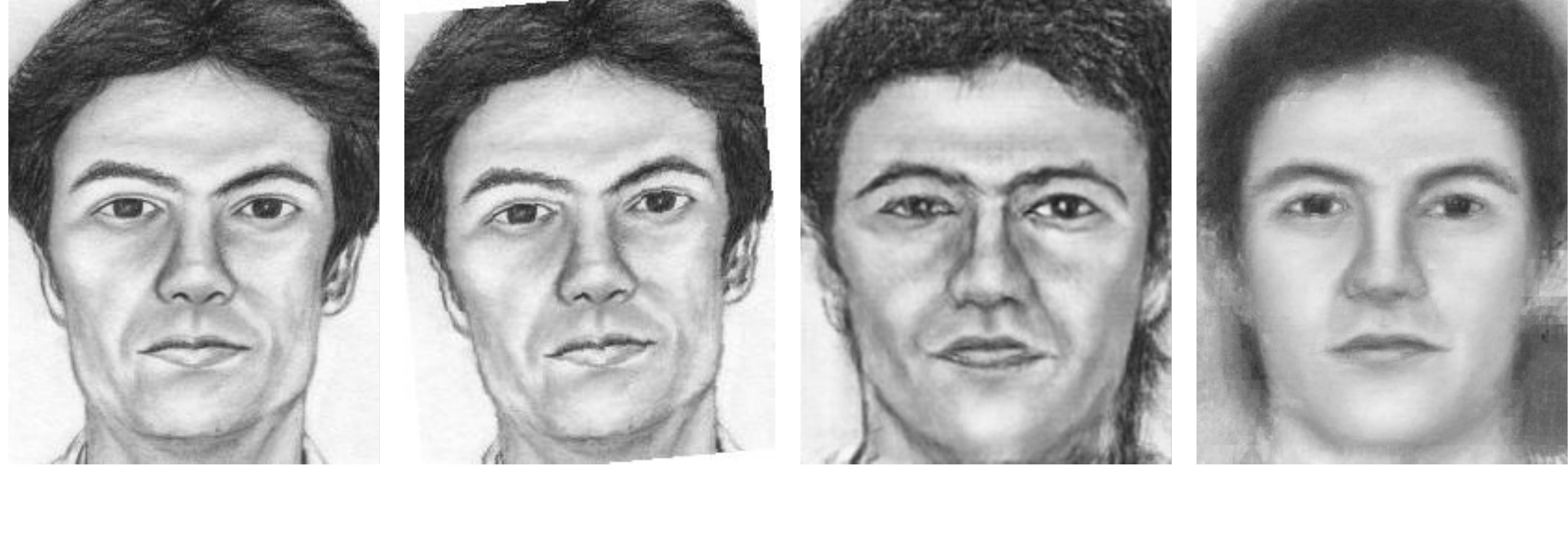}
    \put(1,0.5){(a) Reference}
    \put(24,0.5){(b) R-Reference}
    \put(54.5,0.5){(c) Pix2pix}
    \put(79,0.5){(d) MWF}
    \end{overpic}
    \vspace{-7pt}
    \caption{\small \textbf{Meta-measure 2: Rotation Sensitivity.}
    \vspace{-12pt}
    }\label{fig:MM2}
\end{figure}
}

\noindent\textbf{Meta-measure 1: Stability to Slight Resizing.}
The first meta-measure (MM1) specifies that the rankings of
synthetic sketches should not change much with
slight changes in the reference sketch.
Therefore, we perform a minor 5 pixels downsizing of the reference
by using nearest-neighbor interpolation.
\figref{fig:MM1} gives an example. The hair of the reference
in (b) drawn by the artist has a slight size discrepancy
compared to the photo (a). We observe that about 5 pixels
deviation (\figref{fig:MM1}(c)) in the boundary is common.
Although the two sketches (e) \& (f) are almost identical, widely-used
metrics, \eg, SSIM~\cite{wang2004image}, VIF~\cite{sheikh2006image},
and GMSD~\cite{xue2014gradient} switched the ranking of the two
synthetic sketches (g, h) when using (e) or (f) as the reference.
However, the proposed \ourmeasure~metric consistently ranked (g) higher than (h).

For this meta-measure, we applied the
$\theta=1-\rho$~\cite{best1975algorithm} measure to test the metric
ranking stability before and after the reference downsizing was performed.
The value of $\theta$ falls in the range [0, 2].

\tabref{tab:MetricScore} shows the results: the lower the result is,
the more stable a metric is to slight downsizing.
We can see a significant ($\approx$ 77\% and 83\%) improvement
over the existing SSIM, FSIM, GMSD, and VIF metrics in both the CUFS and CUFSF databases.
These improvements are mainly because the proposed metric considers ``block-level''
statistics rather than ``pixel-level''.

\CheckRmv{
\begin{figure}[t!]
\small
\centering
    \begin{overpic}[width=.69\columnwidth]{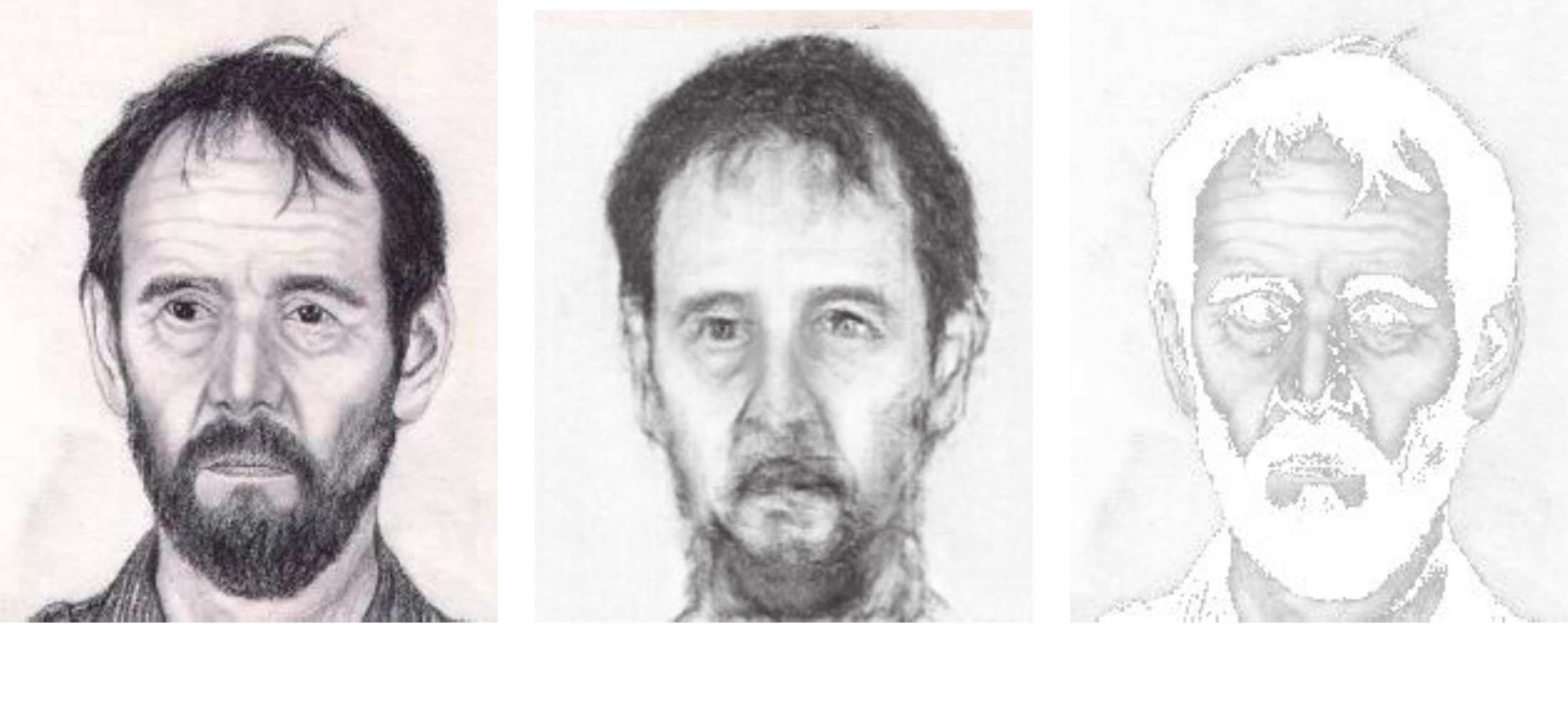}
    \put(1,-1){(a) Reference}
    \put(37,-1){(b) Synthetic}
    \put(75,-1){(c) Light}
    \end{overpic}
    \vspace{-7pt}
    \caption{\small
    \textbf{Meta-measure 3: Content Capture Capability.}
    }\label{fig:MM3}
    \vspace{-12pt}
\end{figure}
}

\CheckRmv{
\begin{table*}[t!]
  \centering
  \small
  \renewcommand{\arraystretch}{1.0}
  \renewcommand{\tabcolsep}{2.8mm}
  \begin{tabular}{rcccccccccccccc}
  \thickhline
  &\multicolumn{3}{c}{\emph{CUFS}~\cite{wang2009face}} & \emph{RCUFS} & \multicolumn{3}{c}{\emph{CUFSF}~\cite{zhang2011coupled}} & \emph{RCUFSF} \\
  \cline{2-4}   \cline{6-8}
   Metic &  MM1$\downarrow$   & MM2$\downarrow$  & MM3$\uparrow$ & Jud$\uparrow$
         &  MM1$\downarrow$   & MM2$\downarrow$  & MM3$\uparrow$ & Jud$\uparrow$ &\\
   &resize   & rotation & content & judgment & resize   & rotation & content & judgment \\
 \thickhline
 \multicolumn{1}{r}{Classical \& Widely Used} &\\
 \toprule
     IFC~\cite{sheikh2005information}
         & 0.256& 0.189& 1.20\% &26.9\%
         & 0.089& 0.112& 3.07\% &25.4\% & \\
     SSIM~\cite{wang2004image}
         & 0.162& 0.086& 81.4\% &37.3\%
         & 0.073& 0.074& 97.4\% &36.8\% & \\
     FSIM~\cite{zhang2011fsim}
         & 0.268& 0.123& 14.2\% &50.0\%
         & 0.151& 0.058& 32.4\% &37.5\% & \\
     VIF~\cite{sheikh2006image}
         & 0.322& 0.236& 43.5\% &44.1\%
         & 0.111& 0.150& 22.2\% &52.8\% & \\
     GMSD~\cite{xue2014gradient}
         & 0.417& 0.210& 21.9\% &42.6\%
         & 0.259& 0.132& 63.6\% &58.6\% & \\
     \rowcolor{mygray}
     \textbf{Scoot (Ours)}
         & \textbf{0.037}& \textbf{0.025}& \textbf{95.9\%} & \textbf{76.3\%}
         & \textbf{0.012}& \textbf{0.008} & \textbf{97.5\%} & \textbf{78.8\%}&\\
  \thickhline
   \multicolumn{1}{r}{Texture-based \& Edge-based}&\\
  \toprule
      Canny~\cite{canny1986computational}
         & 0.086& 0.078& 33.7\% &27.8\%
         &0.138& 0.146& 0.00\%& 0.10\% & \\
      Sobel~\cite{sobel1990isotropic}
         & 0.040& 0.037& 0.00\% &32.8\%
         &0.048& 0.044& 0.00\%& 52.6\% & \\
      GLRLM~\cite{galloway1974texture}
         & 0.111& 0.111& 18.6\% &73.7\%
         &0.125& 0.079& 64.6\%& 68.0\% & \\
      Gabor~\cite{gabor1946theory}
         & 0.062& 0.055& 0.00\% &72.2\%
         &0.089& 0.043& 19.3\%& \textbf{80.9\%} & \\
      \rowcolor{mygray}
      \textbf{Scoot (Ours)}
         & \textbf{0.037}& \textbf{0.025}& \textbf{95.9\%} &\textbf{76.3\%}
         &\textbf{0.012}& \textbf{0.008}& \textbf{97.5\%}& 78.8\% & \\
  \thickhline
   \multicolumn{1}{r}{Feature Combination}&\\
  \toprule
      $\mathcal{HEC}$
         & 0.034& 0.024& 95.9\% & 76.3\% &0.011
         & 0.008& 97.4\%& 78.7\% & \\
      $\mathcal{H}$
         & 0.007& 0.005& 61.5\% &77.5\% &0.003
         & 0.003& 79.1\%& 77.8\% & \\
      $\mathcal{E}$
         & 0.200& 0.104& 98.5\% &73.1\% &0.044
         & 0.026& 99.2\%& 77.4\% & \\
      $\mathcal{C}$
         & 0.010& 0.007& 54.4\% &74.6\% &0.009
         & 0.006& 64.7\%& 73.4\% & \\
      $\mathcal{HC}$
         & 0.011& 0.007& 60.1\% &74.6\% &0.007
         & 0.005& 78.1\%& 73.7\% & \\
      $\mathcal{HE}$
         & 0.156& 0.088& 97.9\% &75.7\% &0.030
         & 0.017& 98.8\%& 80.3\% & \\
      \rowcolor{mygray}
      \textbf{$\mathcal{CE}$ (Scoot)}
         & 0.037& 0.025& 95.9\% &76.3\%&0.012
         & 0.008& 97.5\%& 78.8\% & \\
  \thickhline
  \end{tabular}
  \vspace{-5pt}
  \caption{\small
  \textbf{Benchmark results of classical and alternative texture/edge based metrics.}
  The best result is highlighted in \textbf{bold}, and these differences are all statistically significant at
the $\alpha<0.05$ level. The $\uparrow$ indicates that the higher the score is,
  the better the metric performs, and vice verse ($\downarrow$).
  }\label{tab:MetricScore}
  \vspace{-15pt}
\end{table*}
}

\noindent\textbf{Meta-measure 2: Rotation Sensitivity.}
In real-world situations, sketches drawn by artists may also
have slight rotations compared to the original photographs.
Thus, the proposed second meta-measure (MM2) verifies the sensitivity of reference rotation
for the evaluation metric. We did a slight counter-clockwise rotation ($5^o$)
for each reference.
\figref{fig:MM2} shows an example.
When the reference (a) is switched to the slightly rotated reference (b), the ranking results should not change much.
In MM2, we got the ranking results for each metric
by applying reference sketches and slightly rotated reference sketches (R-Reference) separately.
We utilized the same measure ($\theta$) as meta-measure 1
to evaluate the rotation sensitivity.

The sensitivity results are shown in \tabref{tab:MetricScore}.
It is worth noting that MM2 and MM1 are two aspects of the expected
property described in \secref{sec:Meta-measures}.
Our metric again significantly outperforms the current metrics over the CUFS and CUFSF databases.

\noindent\textbf{Meta-measure 3: Content Capture Capability.}
The third meta-measure (MM3) describes that a good metric
should assign a complete sketch generated by a SOTA algorithm a
higher score than any sketches that only preserve incomplete
strokes.
\figref{fig:MM3} presents an example.
We expect that a metric should prefer the SOTA
synthetic result (b) over the \emph{light strokes}\footnote{
To test the third meta-measure, we use a simple threshold of grayscale
(\eg 170) to separate the sketch (\figref{fig:MM3} Reference) into darker strokes
\& lighter strokes. The image with lighter strokes loses the main
texture features of the face (\eg hair, eye, beard), resulting in an incomplete sketch.
} result (c).
For MM3, we compute the \emph{mean score} of 10 SOTA~\cite{zhang2015end,isola2017image,liu2005nonlinear,wang2017data,
wang2009face,zhou2012markov,wang2018random,song2014real,zhu2017deep,wang2017bayesian}
face sketch synthesis algorithms.
The mean score is robust to situations in which a certain model
generates a poor result. We recorded the number of times the mean score of SOTA
synthetic algorithms is higher than a light stroke's score.

\begin{figure*}[t!]
\small
\centering
    \begin{overpic}[width=\textwidth]{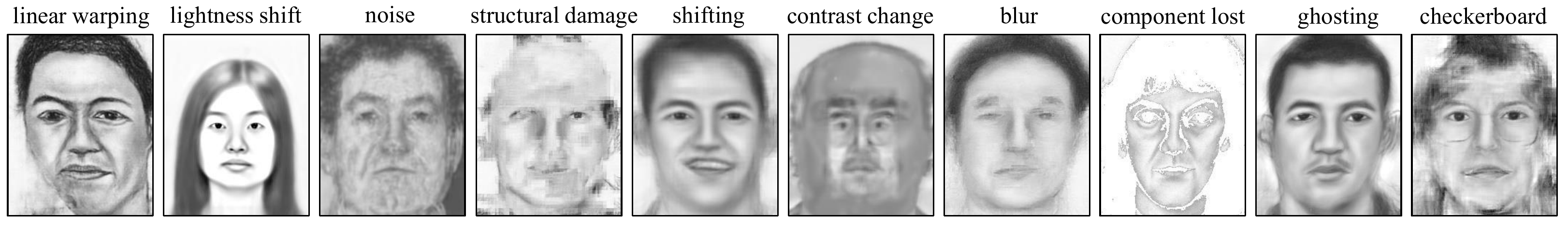}
    \end{overpic}
    \vspace{-20pt}
    \caption{\small \textbf{Our distortions.}
    These distortions are generated by various real synthesis
    algorithms~\cite{zhang2015end,isola2017image,liu2005nonlinear,wang2017data,
    wang2009face,zhou2012markov,wang2018random,song2014real,zhu2017deep,wang2017bayesian}.
    }\label{fig:distortions}
    \vspace{-10pt}
\end{figure*}

For the case shown in \figref{fig:MM3}, the current widely-used
metrics (SSIM, FSIM, VIF) are all in favor of the light sketch. Only the proposed
\ourmeasure~metric gives the correct order.
In terms of pixel-level matching, it is obvious that the regions where dark strokes are
removed are different from the corresponding parts in (a).
But at other positions, the pixels are identical to the reference.
Previous metrics only consider ``pixel-level'' matching and will rank the light strokes
sketch higher. However, the synthetic sketch (b) is better than the light one (c)
in terms of both style and content.
From \tabref{tab:MetricScore}, we observe a great ($\geq$14\%)
improvement over the other metrics in \emph{CUFS} database.
A slight improvement is also achieved for the CUFSF database.

\vspace{-4pt}

\section{Proposed Dataset}
\vspace{-4pt}
Motivated by Zhang \etal~\cite{zhang2018unreasonable}, we also 
established a new perceptual judgments dataset using the 2AFC 
strategy.
These judgments are derived from a wide space of distortions and real algorithm synthesises.
Because the true test of a synthetic sketch assessment metric is on real problems
and real algorithms, we gather perceptual judgments using such outputs.

\vspace{-4pt}
\subsection{Distortions of Facial Sketch}
\vspace{-4pt}

\textbf{Source Images.}
Data on real algorithms is more limited,
as each synthesis model will have its own unique properties.
To obtain more distortion data, we collect
338 pairs (\emph{CUFS}) and 944 pairs
(\emph{CUFSF}) of test set images as source images following the
split scheme of~\cite{wang2018random}.

\textbf{Distortion Types.}
In order to provide more diverse distortion types arise from the real 
synthesis algorithms, in introduce some classical methods~\cite{zhang2015end,isola2017image,liu2005nonlinear,wang2017data,
wang2009face,zhou2012markov,wang2018random,song2014real,zhu2017deep,wang2017bayesian} to achieve this goal.
%
%
As shown in \figref{fig:distortions}, we
introduce 10 distortion types, such as
lightness shift, foreground noise, shifting,
linear warping, structural damage, contrast change,
blur, component lost, ghosting, and
checkerboard artifact.

\vspace{-4pt}
\subsection{Similarity Assessments}
\vspace{-4pt}
\textbf{Data selection.}
21 viewers, who were pre-trained with 50 pairs of ranking, are
asked to rank the synthetic sketch result based on two criteria:
texture similarity and content similarity.
To minimize the ambiguity of human ranking, we follow the voting strategy~\cite{wang2016evaluation}
to conduct this experiment ($\sim$152K judgments) through the following stages:

\begin{itemize}
  \vspace{-5pt}
  \item
    We let the first group of viewers (7 subjects) select four out of ten sketches for each photo.
    The 4 sketches should consist of two good and two bad ones.
    Thus, we are left with 1352 (4 $\times$ 338), and 3776 (4 $\times$ 944)
    sketches for \emph{CUFS} and \emph{CUFSF}, respectively.

  \vspace{-5pt}
  \item
   For the selected four sketches in each photo, the second group
   of viewers (seven people) is further asked to choose three sketches for which they can rank them easily.
   Based on the voting results of viewers, we pick out the 3 most frequently selected sketches.

  \vspace{-5pt}
  \item
  Sketches that are too similar will make it difficult for viewers to judge which
  sketch is better, potentially causing them to give random decisions.
  To avoid this random selection, we ask the last group of viewers (seven)
  to pick out the pair of sketches that are most obvious to rank.
  \vspace{-6pt}
\end{itemize}

\textbf{2AFC strategy~\cite{zhang2018unreasonable}.}
For each image, we have a reference sketch $r$ drawn by artists and
two distortions $s_0, s_1$. We ask the viewer which is closer
to the reference $s$, and the answer limit to $q\in \{0,1\}$.
The average judgment time per image is about 2 seconds.
%
Note that we have 5 volunteers involved in the whole process
for cross-checking the ranking. For example, if there $\geq4$ viewer
preferences for $s_0$ and $\leq1$ for $s_1$, the final ranking will be
$s_0>s_1 \& q=1$. All triplets with a clear majority will be preserved and the other triplets are discarded.
Finally, we establish two new human-ranked\footnote{
The two datasets include 1014 (3$\times$338 triplets), and 2832 (3$\times$944 triplets)
human-ranked images, respectively.
Recent works~\cite{sun2016benchmark,wu2019ip102} show that the scale of
a dataset is important.
To our best knowledge, this is the first large-scale
publicly available human judgment dataset in FSS.
} datasets:
\emph{\textbf{RCUFS}} and \emph{\textbf{RCUFSF}}.
Please refer to our \supp{\href{http://dpfan.net/Scoot/}{website}} for complete datasets.

\vspace{-4pt}
\subsection{Human Judgments}
\vspace{-4pt}
Here, we evaluate how well our Scoot and other
compared metrics. The \emph{RCUFS} and \emph{RCUFSF}
contain 338 and 944 judged triplets, respectively.
To increase an inherently noisy process, we compute the agreement
of a metric with each triplet and adopt the \emph{average} statistics among
the dataset as the final performance.

\textbf{How well do classical metrics and our Scoot perform?}
\tabref{tab:MetricScore} shows the performance of various
classical metrics (\eg, IFC, SSIM, FSIM, VIF, and GMSD).
Interestingly, these metrics perform at about the same low level (\eg, $\leq$59\%).
The main reason is that these metrics were not originally designed for pixel mismatching which is common in FSS.
%
However, the proposed Scoot metric shows a significant ($\sim$26.3\%)
improvement over the best prior metric in \emph{RCUFS}.
This improvement is due to our consideration of structure and texture similarity
which human perception considers as two essential factors
when evaluating sketches.

\vspace{-4pt}
\section{Discussion}\label{sec:AblationStudy}
\vspace{-4pt}

\textbf{Which elements are critical for their success?}
In \secref{sec:co-occurrenceTexture},
we considered 3 widely-used statistics:
Homogeneity ($\mathcal{H}$), Contrast ($\mathcal{C}$), and Energy ($\mathcal{E}$).
To achieve the best performance, we need to explore the
best combination of these statistical features.
We have applied our three meta-measures as well as human judgments
to test the performance of the \ourmeasure~metric using each single feature,
each feature pair and the combination of all three features.

\CheckRmv{
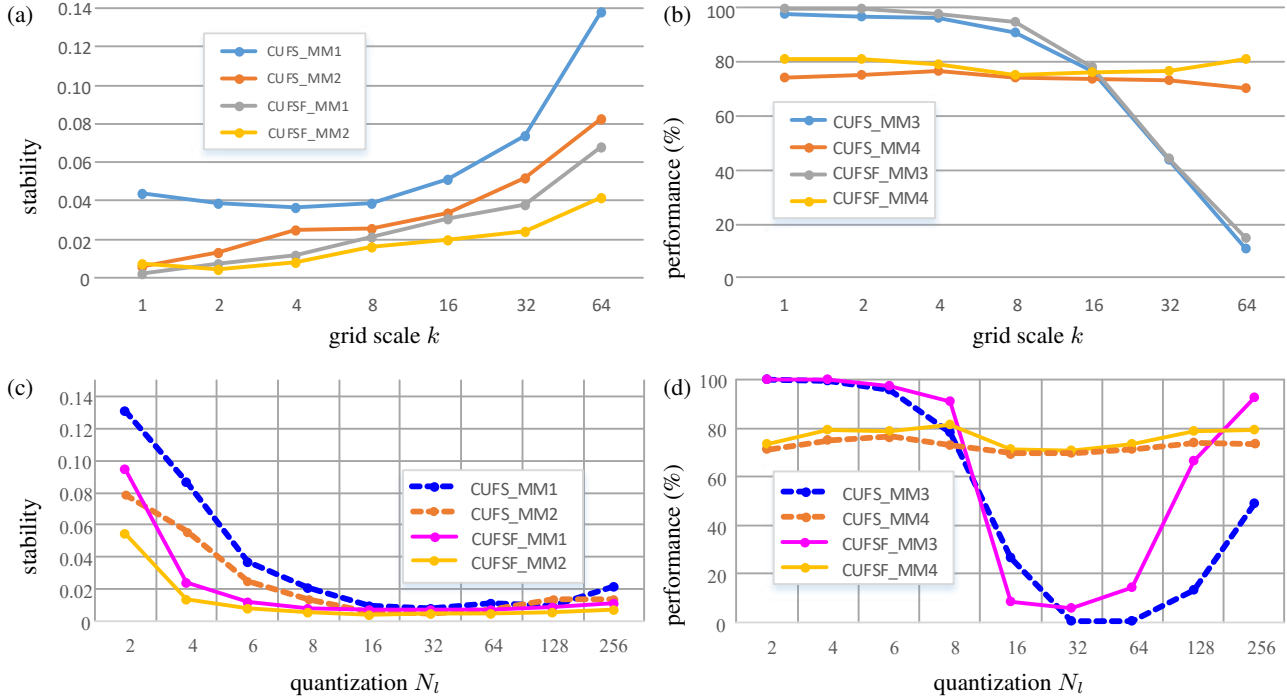
\begin{figure*}[t!]
\small
\centering
    \begin{overpic}[width=.98\textwidth]{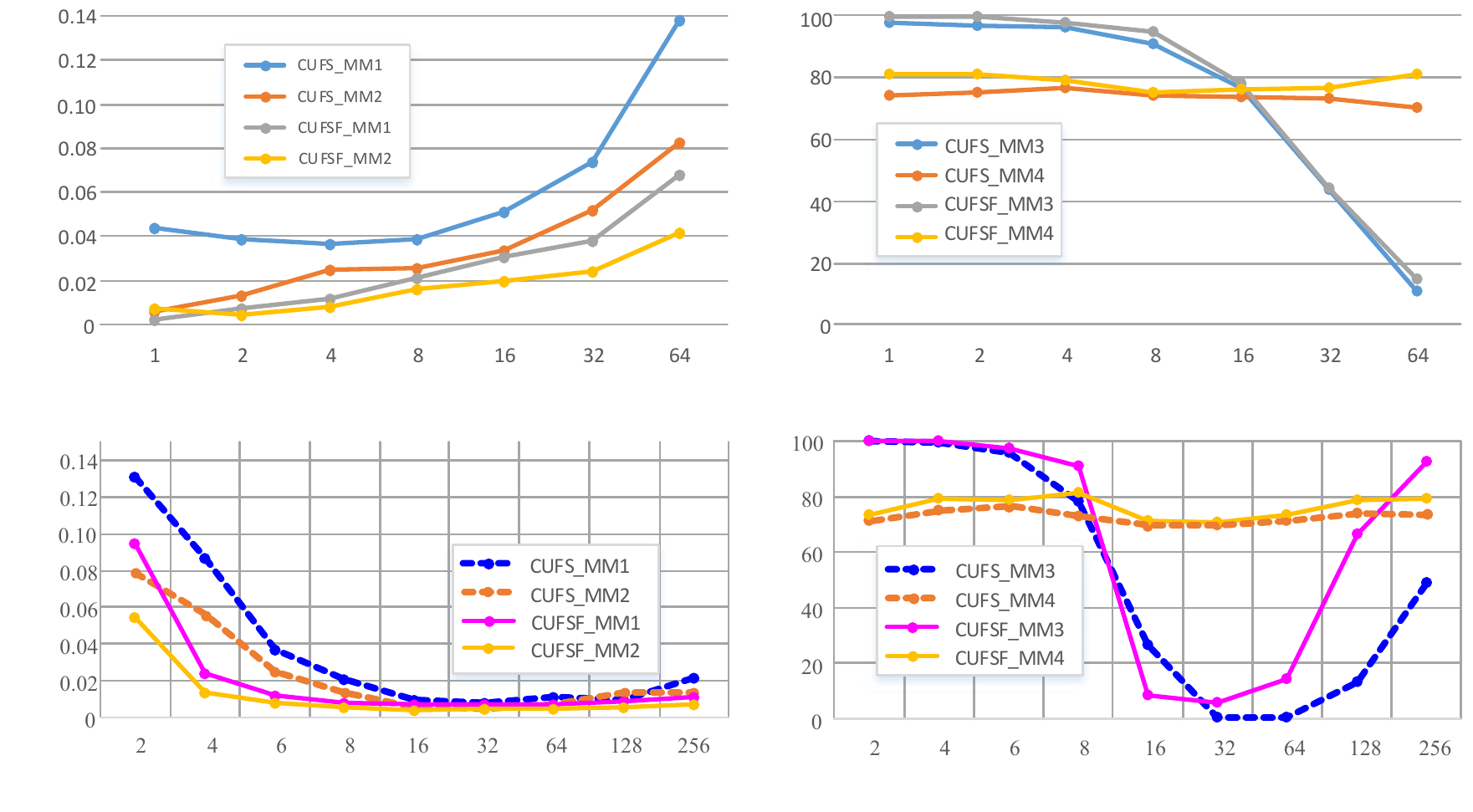}
    \put(0,53){(a)}
    \put(51,53){(b)}
    \put(0,24){(c)}
    \put(51,24){(d)}
    \put(1,38){\rotatebox{90}{\small stability}}
    \put(51,33){\rotatebox{90}{\small performance (\%)}}

    \put(1,10){\rotatebox{90}{\small stability}}
    \put(51,6){\rotatebox{90}{\small performance (\%)}}

    \put(25,28){\small grid scale $k$}
    \put(75,28){\small grid scale $k$}

    \put(22,1){\small quantization $N_l$}
    \put(72,1){\small quantization $N_l$}
    \end{overpic}
    \vspace{-6pt}
    \caption{\small
    \textbf{Sensitivity experiments of the spatial structure (top) and quantization (bottom)}.
    For MM1 \& MM2, the lower the better.
    For MM3 \& MM4, the higher the better.
    }\label{fig:Sensitive}
    \vspace{-10pt}
\end{figure*}
}

The results are shown in~\tabref{tab:MetricScore}.
All possibilities ($\mathcal{H}$, $\mathcal{E}$, $\mathcal{C}$, $\mathcal{CE}$,
$\mathcal{HE}$, $\mathcal{CH}$, $\mathcal{HEC}$) perform well in Jud (human judgment).
$\mathcal{H}$ and $\mathcal{C}$ are insensitive to
re-sizing (MM1) and rotation (MM2), while they are not good
at content capture (MM3). $\mathcal{E}$ is the opposite compared to $\mathcal{H}$ and $\mathcal{C}$.
Thus, using a single feature is not good.
The results of combining two features show that if $\mathcal{H}$ is combined with $\mathcal{E}$,
the sensitivity to re-sizing and rotating will still be high,
while partially overcoming the weakness of $\mathcal{E}$.
The performance of $\mathcal{H+E+C}$ shows no improvement
compared to the combination of ``$\mathcal{CE}$'' features.
Previous work in~\cite{baraldi1995investigation} also found the energy and contrast
to be the most efficient features for discriminating textural patterns.
Thus, we choose ``$\mathcal{CE}$'' feature as our final combination to
extract the perceptual features.

\textbf{How well do these ``perceptual features'' actually correspond to human visual
perceptions?}
As described in \secref{sec:co-occurrenceTexture}, sketches are quite close to textures.
There are many other texture \& edge-based features (\eg GLRLM~\cite{galloway1974texture},
Gabor~\cite{gabor1946theory}, Canny~\cite{canny1986computational}, Sobel~\cite{sobel1990isotropic}).
Here, we select the most wide-used features as candidate alternatives to replace our ``$\mathcal{CE}$'' feature.
For GLRLM, we select all five statistics mentioned in the original version.
Results are shown in \tabref{tab:MetricScore}.
Gabor and GLRLM are texture features, while the other two are edge-based.
All the texture features (GLRLM, Gabor) and the proposed \ourmeasure~metric
provide a good (\eg, $\geq$68\%) consistency with human ranking (Jud).
Among all the texture features, the proposed metric provides a
consistently high average performance with human ranking (Jud).
GLRLM performs well according to MM1 \& 2 \& 3.
Gabor is reasonable in terms of MM1 \& 2, but not good at MM3.
For edge-based features, Canny fails according to all meta-measures.
Sobel is very stable to slight re-sizing (MM1) or rotating (MM2),
but cannot capture content (MM3) and is not consistent with human judgment (Jud).
Interestingly, Canny, Sobel, and Gabor assigned the incomplete stroke a higher score than
the sketch generated by the SOTA algorithm. In other words, the metric has completely
reversed the ranking results for all the tested cases.
In terms of overall results, we conclude that our ``$\mathcal{CE}$'' feature is more robust than
other competitors.

\noindent\textbf{What is the Sensitivity to the Spatial Structure?}
To analyze the effect of spatial structure, we derive seven
variants, each of which divides the sketch with a different sized grid, \ie, $k$ is
set to 1, 2, 4, $\cdots$, 64.
The results of MM3 \& 4 in \figref{fig:Sensitive}(b)
show that $k$ = 1 achieves the best performance.
However, the weakness of this version is that it only captures the
``image-level'' statistics, and the structure of the sketch is ignored.
That is, a sketch made up of an arbitrary arrangement can also achieve a high score.
%
The experiment of MM1 in \figref{fig:Sensitive}(a),
clearly shows that
$k$ = 4 achieves the best performance for the CUFS dataset.
Based on the two experiments, $k=4$ gains the most robust performance.

\noindent\textbf{What is the Sensitivity to Quantization?}
To determine which quantization parameter $N_l$ (baseline: $N_l$ = \{2, 4, 6, 8, 16, 32, 64, 128\})
produces the best performance we perform a further sensitivity test.
From \figref{fig:Sensitive}(c)\&(d),
we observe that quantizing the input sketch to 32 grey levels achieves
an excellent result. However, for the experiments of MM3 \& MM4, it gains the worst performance.
Considering overall experiments, $N_l=6$ achieves a more robust result.

\vspace{-4pt}
\section{Conclusion}\label{sec:otherFeature}
\vspace{-4pt}
In this work, we explore the human perception problem,
\eg, what is the difference between human choice and metrics.
A tool used to analyze the above question are facial sketches.
We provide a specific metric, called Scoot
(Structure Co-Occurrence Texture), that captures
human perception, and is analyzed by the proposed three meta-measures.
Finally, we built the first human-perception-based sketch database
that can evaluate how well a metric is in line with human perception.
We systematically evaluate different texture-based/edge-based features on our
Scoot architecture and compare them with classic metrics.
Our results show that ``spatial structure'' and ``co-occurrence'' texture
are two generally applicable perceptual features in facial sketches.
%
In the future, we will continue to develop and apply Scoot in order to
further push the frontiers of research, \eg, for evaluation of background subtraction~\cite{yizhe_sub_2017}.

\vspace{-15pt}
\paragraph{Acknowledgments.}
This research was supported by NSFC (61572264, 61620106008, 61802324, 61772443),
the national youth talent support program, and Tianjin Natural Science
Foundation (17JCJQJC43700, 18ZXZNGX00110).


{\small
\bibliographystyle{ieee_fullname}
\bibliography{SketchMeasure}
}

\end{document}